\documentclass[10pt,twocolumn,letterpaper]{article}

\usepackage[pagenumbers]{cvpr} 
\usepackage{multirow}   
\definecolor{cvprblue}{rgb}{0.21,0.49,0.74}
\usepackage[pagebackref,breaklinks,colorlinks,allcolors=cvprblue]{hyperref}

\title{UNIC-Adapter: Unified Image-instruction Adapter with Multi-modal \\ Transformer for Image Generation}

\author{
Lunhao Duan\footnotemark[1]~~\textsuperscript{1}\quad 
Shanshan Zhao\footnotemark[2]~~\textsuperscript{2}\quad
Wenjun Yan\textsuperscript{2}\quad
Yinglun Li\textsuperscript{2}\quad
Qing-Guo Chen\textsuperscript{2}\\
Zhao Xu\textsuperscript{2}\quad
Weihua Luo\textsuperscript{2}\quad
Kaifu Zhang\textsuperscript{2}\quad
Mingming Gong\textsuperscript{3}\quad
Gui-Song Xia\footnotemark[2]~~\textsuperscript{1}\\ \\
\textsuperscript{1}School of Computer Science, Wuhan University \\
\textsuperscript{2}Alibaba International Digital Commerce\quad 
\textsuperscript{3}University of Melbourne
}

\begin{document}
\maketitle
\renewcommand{\thefootnote}{\fnsymbol{footnote}}
\footnotetext[1]{This work was done when Lunhao Duan was a research intern at Alibaba International Digital Commerce.} 
\footnotetext[2]{Corresponding authors.}
\begin{abstract}
Recently, text-to-image generation models have achieved remarkable advancements, particularly with diffusion models facilitating high-quality image synthesis from textual descriptions.
However, these models often struggle with achieving precise control over pixel-level layouts, object appearances, and global styles when using text prompts alone. 
To mitigate this issue, previous works introduce conditional images as auxiliary inputs for image generation, enhancing control but typically necessitating specialized models tailored to different types of reference inputs.
In this paper, we explore a new approach to unify controllable generation within a single framework.
Specifically, we propose the unified image-instruction adapter (UNIC-Adapter) built on the Multi-Modal-Diffusion Transformer architecture, to enable flexible and controllable generation across diverse conditions without the need for multiple specialized models. 
Our UNIC-Adapter effectively extracts multi-modal instruction information by incorporating both conditional images and task instructions, injecting this information into the image generation process through a cross-attention mechanism enhanced by Rotary Position Embedding.
Experimental results across a variety of tasks, including pixel-level spatial control, subject-driven image generation, and style-image-based image synthesis, demonstrate the effectiveness of our UNIC-Adapter in unified controllable image generation.
\end{abstract}
\vspace{-4mm}    
\begin{figure}[t]
  \centering
   \includegraphics[width=0.85\linewidth]{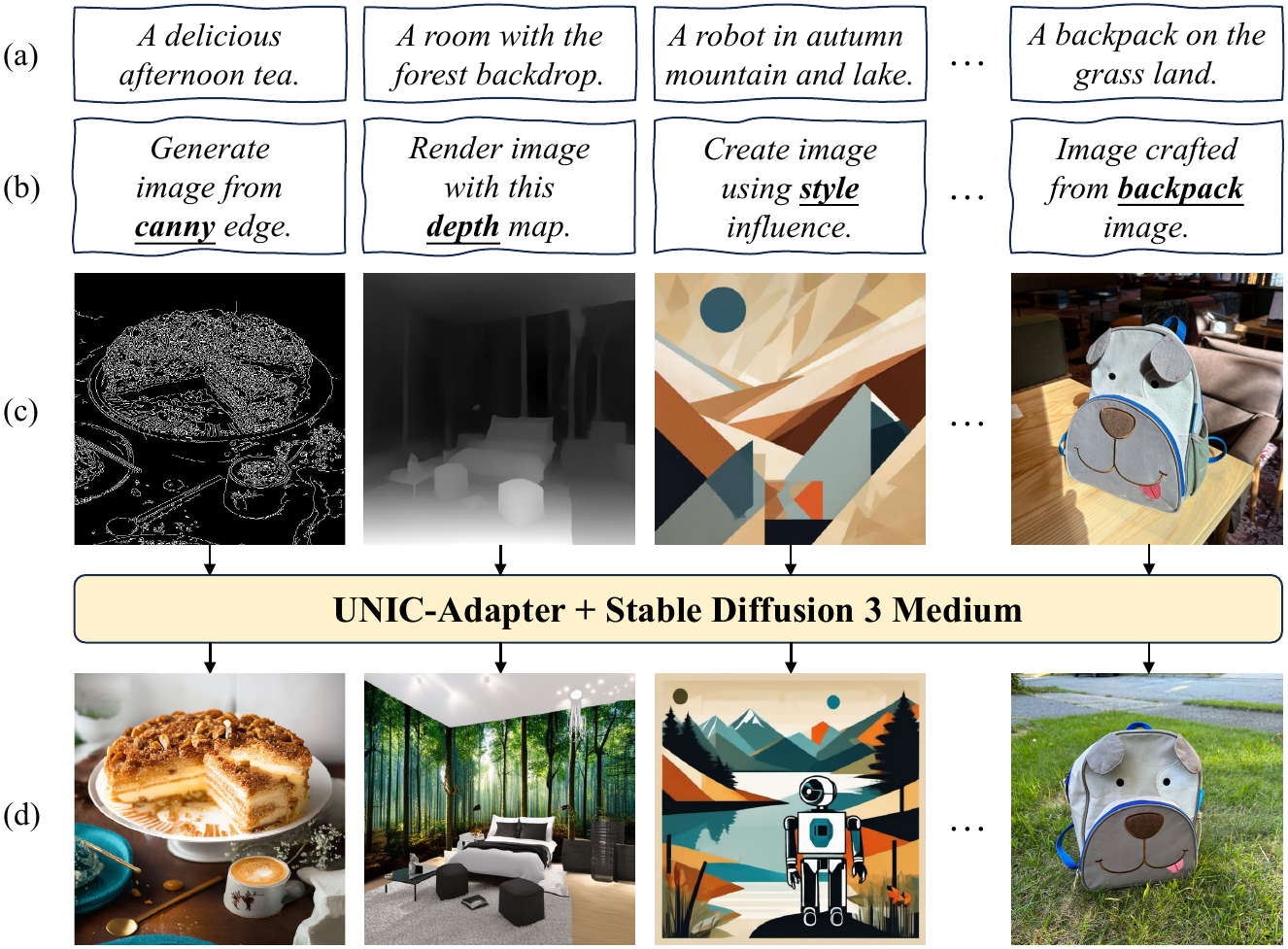}
   \caption{With the UNIC-Adapter, SD3 enables flexible and controllable generation across multiple reference modalities within a single model. (a) to (d) represent text prompts, task instructions, conditional images, and generated images, respectively.}
   \vspace{-4mm}
   \label{fig:teaser}
\end{figure}

\section{Introduction}
\label{sec:intro}
Recent advancements in text-to-image (T2I) generation models~\cite{LDM,podell2023sdxl,imagen3,dalle3,chen2023pixart,sd3,liu2024playground,kolors,blackforestlabs_flux_2024}, primarily driven by diffusion models~\cite{ddpm}, have been witnessed in recent years. By leveraging large-scale paired text-image data~\cite{schuhmann2022laion}, numerous open-source T2I models, like Stable Diffusion 1.5 (SD1.5)~\cite{LDM}, Stable Diffusion XL (SDXL)~\cite{podell2023sdxl}, Pixart-series~\cite{chen2023pixart,chen2024pixart_d, chen2024pixart_s}, Stable Diffusion 3 (SD3)~\cite{sd3}, Hunyuan-DiT~\cite{li2024hunyuan}, and FLUX.1-dev~\cite{blackforestlabs_flux_2024} have
significantly enhanced the fidelity of images generated from natural language descriptions.
These models utilize a range of architectures, spanning from U-Net-based designs~\cite{unet} to the more recent Diffusion-Transformer-based frameworks~\cite{dit}.

Despite these advancements, relying solely on text prompts often falls short in defining pixel-level spatial structures and geometric details, exact object appearances, and global styles of a specific image. 
To address this limitation, recent research~\cite{controlnet,ye2023ipadapter,t2iadapter,pan2023kosmos-g,wang2024instantid,gao2024styleshot,ruiz2023dreambooth,huang2023composer} has incorporated conditional images as additional inputs alongside text prompts. 
These conditional images offer finer control over pixel-level~\cite{controlnet,zheng2023layoutdiffusion}, object-level~\cite{ye2023ipadapter,pan2023kosmos-g,wang2024instantid}, or style-specific~\cite{gao2024styleshot,hertz2024style,wang2024instantstyle} features in the generated outputs.
For example, ControlNet~\cite{controlnet} enables pixel-level spatial control by integrating structural inputs, such as depth and edge maps, through a parallel encoder tailored for each type of control. 
Similarly, IP-Adapter~\cite{ye2023ipadapter} enables content-specific control by conditioning on a reference image and injecting its CLIP-based~\cite{clip} embeddings via cross-attention layers. 
However, these methods often necessitate individual model training for each control type, resulting in increased training costs and complicating the integration of different control inputs. 
IPAdapter-Instruct~\cite{ipadapter_instruct} extends IP-Adapter by incorporating instruction prompts to specify task types for different conditional images, enabling multi-task handling within a single model. 
UniControl~\cite{qin2023unicontrol} tackles various pixel-level control tasks within a unified model by introducing task-specific adapters to extract features for different visual conditions. 
Meanwhile, Uni-ControlNet~\cite{zhao2024uni-controlnet} divides conditional images into two groups: local (pixel-level) and global (image-level) conditions, and designs dedicated adapters for each group.
However, this approach is not fully unified, as the two adapters are trained separately.

To further unify diverse reference inputs, including text, edge maps, content images, style images, etc., Instruct-Imagen~\cite{instruct_imagen} represents them as multi-modal instructions within a single framework.
OmniGen~\cite{xiao2024omnigen}, a latest work, proposes using a transformer model~\cite{abdin2024phi} for various image generation tasks with multi-modal interleaved text and images as inputs.
To adapt the model for multi-modal instructions, these two methods involve multi-stage training of the entire model.
To enable more efficient training for unified controllable image generation, this paper seeks to design an adapter based on a pre-trained T2I model, avoiding the need to train the entire model.

Given that DiT-based architectures~\cite{peebles2023scalable} equipped with pure transformer blocks have become the mainstream in T2I models~\cite{chen2023pixart,li2024hunyuan,sd3,blackforestlabs_flux_2024}, we choose SD3, an open-source state-of-the-art (SOTA) model, as the foundation model for studying the adapter. 
To achieve unified controllable image generation across various conditions, we employ the Multi-Modal-Diffusion Transformer (MM-DiT) structure~\cite{sd3,blackforestlabs_flux_2024}, which enables full cross-attention between text and image features to extract multi-modal instruction information.
By treating text as an independent modality, the MM-DiT allocates substantial model capacity to capture and understand text-based information, thereby improving its interpretation of text prompts~\cite{sd3}. 
Leveraging the strengths of the MM-DiT, we propose a unified image-instruction adapter (UNIC-Adapter) to facilitate image generation conditioned on various types of reference images with a task instruction indicating the desired generation target.
Specifically, task instruction and conditional image features are extracted by text encoders~\cite{clip,t5} and a Variational Autoencoder (VAE)~\cite{sd3}, respectively.
Then, the adapter processes instruction and image features using MM-DiT blocks, which are subsequently integrated into the main generation branch via a cross-attention mechanism. 
To further enhance spatial awareness, which is essential for pixel-level control, we incorporate Rotary Position Embedding (RoPE)~\cite{rope} into the query and key features within the cross-attention. With the proposed UNIC-Adapter, SD3 can generate images under a wide array of conditions, as shown in Figure~\ref{fig:teaser}.

In summary, our contributions are threefold:
\begin{itemize}[leftmargin=*]
    \item We are the first to leverage the MM-DiT architecture for unified controllable generation across various types of reference images in T2I models.
    \item We introduce an effective feature injection method based on cross-attention, enhanced with RoPE encoding for better spatial control.
    \item Equipped with our UNIC-Adapter, SD3 enables unified controllable generation across 14 types of conditional images in our experiments.
\end{itemize}

\section{Related Work}
\label{sec:related}
\subsection{Text-to-Image Generation Models} 
Mainstream T2I generation models~\cite{LDM,podell2023sdxl,chen2023pixart,sd3} are primarily based on diffusion models~\cite{ddpm}, which generate images by learning to reverse a progressive noise-adding process. 
In these models, text input typically serves as a conditional signal, guiding the generation process to align with the provided description. 
The success of transformer architectures~\cite{vaswani2017attention}, particularly in large language models~\cite{touvron2023llama,openai2024gpt4technicalreport}, has significantly influenced T2I model architectures, evolving them from convolution-transformer hybrids~\cite{LDM,podell2023sdxl} to fully transformer-based designs~\cite{chen2023pixart,chen2024pixart_d,chen2024pixart_s,sd3,li2024hunyuan,blackforestlabs_flux_2024}.
Recently, the MM-DiT architecture~\cite{sd3,blackforestlabs_flux_2024} has advanced this evolution by treating text and image features as distinct modalities with full-attention mechanisms, enabling each modality to attend to the other.
This approach outperforms traditional cross-attention setups, where text features are used only as keys and values~\cite{sd3}.
However, relying solely on text-based conditioning presents limitations, particularly for images that require detailed spatial layouts, specific object appearances, or style control.
Our work builds upon these transformer-based diffusion models by integrating additional conditional images and task instructions, thereby enhancing the flexibility and controllability of T2I generation. 

\begin{figure*}[t]
  \centering
   \includegraphics[width=0.85\linewidth]{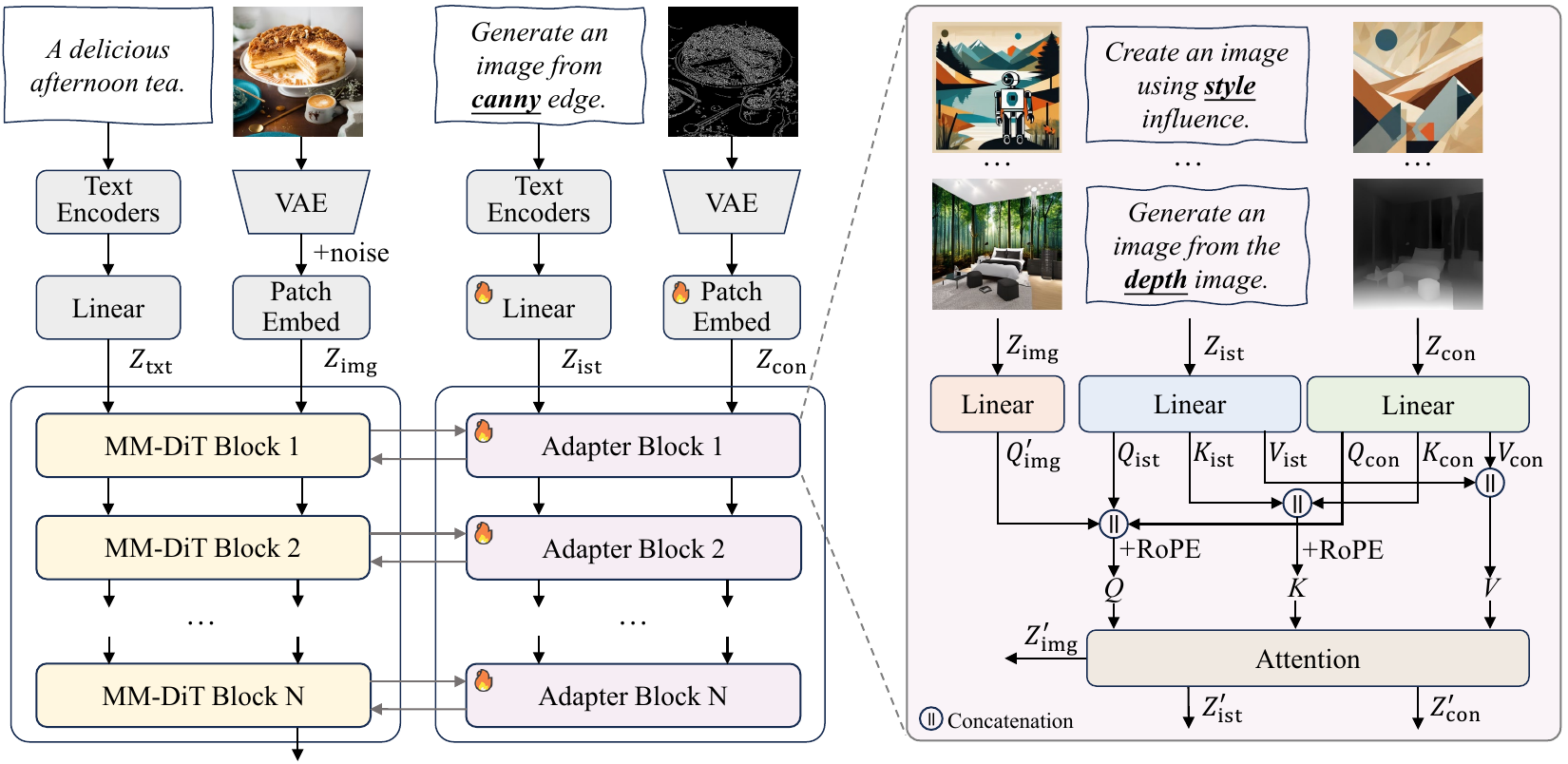}
   \caption{The overall architecture of our proposed UNIC-Adapter. The task instruction and conditional image features are progressively attending to each other through a series of $N$ adapter blocks. In each adapter block, the image features $Z_{\text{img}}$ from MM-DiT block in the main image generation branch serve as the query, while both task instruction features $Z_{\text{ist}}$ and conditional image features $Z_{\text{con}}$ function as keys and values. For simplicity, normalization layers and feed-forward networks are omitted in this figure.}
   \label{fig:method}
\end{figure*}

\subsection{Conditional Image-based T2I Models}
Conditional image-based T2I models introduce auxiliary image inputs to complement textual prompts, enabling finer control over aspects such as spatial layout~\cite{controlnet,zheng2023layoutdiffusion,controlnetplus}, object content~\cite{ye2023ipadapter,li2024blip_diff,suti,ma2024subject,song2024moma}, and stylistic elements~\cite{gao2024styleshot,han2024stylebooth,wang2024instantstyle}. 
For pixel-level spatial control, ControlNet~\cite{controlnet} employs a duplicated U-Net encoder to process specific types of conditional images (e.g., edge, pose, and depth maps) and integrates the features into the main T2I model through residual connections in the U-Net decoder. 
Another key work, IP-Adapter~\cite{ye2023ipadapter}, uses the Vision Transformer (ViT)~\cite{vit} encoder from CLIP~\cite{clip} to extract features from conditional images. 
These features are then incorporated into the U-Net backbone via cross-attention layers, achieving controllable generation with contents from conditional images.
IPAdapter-Instruct~\cite{ipadapter_instruct} extends this by introducing an instructional prompt that guides the interpretation of conditional images. 
To minimize the necessity for individual model training for each type of conditional image in ControlNet, UniControl~\cite{qin2023unicontrol} unifies nine pixel-level control tasks within a single model, leveraging task-specific feature extraction adapters and a task-aware HyperNet to modulate features based on task instructions. To achieve control with image content as well as spatial layout,
Uni-ControlNet~\cite{zhao2024uni-controlnet} incorporates two dedicated adapters for local (pixel-level control) and global (CLIP-image embedding control) conditions, respectively. 
Unlike these methods, which freeze the base model and only training the newly added modules / adapters, 
Instruct-Imagen~\cite{instruct_imagen} proposes using multi-modal instructions and optimize the entire model, including the base model (Imagen~\cite{saharia2022photorealistic}) and the new cross-attention layers, for image generation under different conditions. A recent work,
OmniGen~\cite{xiao2024omnigen}, develops a unified image generation model for various multi-modal inputs using a transformer-based architecture~\cite{abdin2024phi}.
In this paper, we aim to enhance the generation capabilities of an existing T2I model for diverse conditions by introducing a unified image-instruction adapter, named UNIC-Adapter, without the need to train the entire model. 
\vspace{-2mm}

\section{Method}

In this section, we first provide a brief overview of the DiT and MM-DiT employed in SOTA T2I models. 
Then, we describe our UNIC-Adapter, which incorporates task instructions and conditional images to enhance the model’s capability to interpret various user-specified conditions.

\subsection{Preliminary}

\textbf{Diffusion Transformer Block.} 
In DiT Block-based T2I models~\cite{chen2023pixart}, conditional text features \( Z_{\text{txt}} \) extracted from text encoders~\cite{clip,t5} are integrated into DiT Blocks via cross-attention layers. 
Specifically, in each DiT Block, the conditional text features are first mapped to key \( K_{\text{txt}} \) and value \( V_{\text{txt}} \) features through linear transformations \( {L} \), while the image feature \( Z_{\text{img}} \) is mapped to a query feature \( Q_{\text{img}} \). 
An attention mechanism then updates the image features \( Z_{\text{img}} \) by attending to \( K_{\text{txt}} \) and \( V_{\text{txt}} \):
\begin{small}
\begin{equation}
\begin{aligned}
    & K_{\text{txt}} = {L}_{\text{txt}}^k(Z_{\text{txt}}), V_{\text{txt}} = {L}_{\text{txt}}^v(Z_{\text{txt}}), Q_{\text{img}} = {L}_{\text{img}}^q(Z_{\text{img}}), \\
    & Z_{\text{img}}^{'} = \text{Attn}(Q_{\text{img}}, K_{\text{txt}}, V_{\text{txt}}),
\end{aligned}
\end{equation}
\end{small}where \( {\text{Attn}} \) indicates the multi-head attention~\cite{vaswani2017attention}.
In this setup, the same \( Z_{\text{txt}} \) features are provided to DiT Blocks across different layers as static conditioning signals, which may limit the utilization of text features.

\textbf{Multi-modal Diffusion Transformer Block.}
Recent T2I models~\cite{sd3,blackforestlabs_flux_2024} based on MM-DiT Blocks enhance DiT Blocks by treating conditional text as an independent modality capable of dynamically interacting with image features across layers. 
In MM-DiT, both text and image features serve as queries, keys, and values, enabling cross-modal attention in both directions through a full-attention mechanism. 
Specifically, the MM-DiT layer performs the following operations:
\begin{small}
\begin{equation}
\begin{aligned}
& Q_{\text{txt}} = {L}_{\text{txt}}^q(Z_{\text{txt}}), K_{\text{txt}} = {L}_{\text{txt}}^k(Z_{\text{txt}}), V_{\text{txt}} = {L}_{\text{txt}}^v(Z_{\text{txt}}), \\
& Q_{\text{img}} = {L}_{\text{img}}^q(Z_{\text{img}}), K_{\text{img}} = {L}_{\text{img}}^k(Z_{\text{img}}), V_{\text{img}} = {L}_{\text{img}}^v(Z_{\text{img}}).
\end{aligned}
\end{equation}
\end{small}These queries, keys, and values are processed in a full-attention layer, where each modality attends to both text and image features:
\begin{equation}
\begin{aligned}
& Z_{\text{txt}}^{'} = \text{Attn}(Q_{\text{txt}}, [K_{\text{txt}} \Vert K_{\text{img}}], [V_{\text{txt}} \Vert V_{\text{img}}]), \\
& Z_{\text{img}}^{'} = \text{Attn}(Q_{\text{img}}, [K_{\text{txt}} \Vert K_{\text{img}}], [V_{\text{txt}} \Vert V_{\text{img}}]),
\end{aligned}
\label{eqa:attn_txt_img}
\end{equation}where \( \Vert \) denotes concatenation.  
This enhanced interaction facilitates richer cross-modal feature extraction, as text and image features are iteratively refined across layers, improving text-guided image generation~\cite{sd3}. 
This design has been adopted in recent SOTA T2I~\cite{sd3,blackforestlabs_flux_2024} models.

\subsection{UNIC-Adapter}
As shown in Figure~\ref{fig:method}, the UNIC-Adapter introduces two additional inputs: task instructions and conditional images.
This integration enables unified controllable image generation by allowing the model to interpret and apply specific features from a variety of conditional images, guided by the task instruction. 
Below, we provide a detailed description of its structure.

\textbf{Image-instruction Feature Extraction.}
To enable a comprehensive understanding of task instructions and conditional images, we leverage MM-DiT blocks to enhance the interaction between these inputs.
Specifically, we adopt SD3 as our base model. 
Following the setup in SD3, task instruction features, denoted as \( Z_{\text{ist}} \), are extracted using text encoders such as CLIP-G/14~\cite{clip-g}, CLIP-L/15~\cite{clip}, and T5 XXL~\cite{t5}.
Simultaneously, conditional image features, \( Z_{\text{con}} \), which represent one of several condition types (e.g., edge maps, depth maps, subject content images, or style-reference images), are obtained through the VAE. 
Within the MM-DiT blocks, task instruction and conditional image features undergo mutual attention updates, defined as follows:
\begin{small}
\begin{equation}
\begin{aligned}
& Q_{\text{ist}} = {L}_{\text{ist}}^q(Z_{\text{ist}}), K_{\text{ist}} = {L}_{\text{ist}}^k(Z_{\text{ist}}), V_{\text{ist}} = {L}_{\text{ist}}^v(Z_{\text{ist}}), \\
& Q_{\text{con}} = {L}_{\text{con}}^q(Z_{\text{con}}), K_{\text{con}} = {L}_{\text{con}}^k(Z_{\text{con}}), V_{\text{con}} = {L}_{\text{con}}^v(Z_{\text{con}}), \\
& Z_{\text{ist}}^{'} = \text{Attn}(Q_{\text{ist}}, [K_{\text{ist}} \Vert K_{\text{con}}], [V_{\text{ist}} \Vert V_{\text{con}}]), \\
& Z_{\text{con}}^{'} = \text{Attn}(Q_{\text{con}}, [K_{\text{ist}} \Vert K_{\text{con}}], [V_{\text{ist}} \Vert V_{\text{con}}]).
\end{aligned}
\label{eqa:attn_ist_con}
\end{equation}
\end{small}These updates ensure that the task instruction \( Z_{\text{ist}} \) and conditional image \( Z_{\text{con}} \) dynamically refine each other’s features by selectively attending to relevant information. 

\textbf{Image-instruction Feature Injection.}
To improve the model’s responsiveness to diverse user-defined conditions, the key and value features from both instruction and conditional image inputs are integrated simultaneously into the main image generation branch through cross-attention~\cite{ye2023ipadapter,instruct_imagen,wang2024instantid}.
This mechanism ensures that the feature injection is directed by the specific task instruction.
Experiments indicate that injecting either conditional image features or task instruction features alone through cross-attention results in suboptimal performance, as demonstrated in the ablation studies of Section~\ref{sec:experiment}. 
By contrast, the simultaneous incorporation of both feature types optimally supports unified controlled image generation. 
Accordingly, the image features \( Z_{\text{img}}^{'} \) from Equation~\ref{eqa:attn_txt_img} are further updated as follows:
\begin{equation}
\begin{aligned}
& Q_{\text{img}}^{'} = {L}_{\text{cross}}^q(Z_{\text{img}}), \\
& Z_{\text{img}}^{''} = Z_{\text{img}}^{'} + \text{Attn}(Q_{\text{img}}^{'}, [K_{\text{ist}} \Vert K_{\text{con}}], [V_{\text{ist}} \Vert V_{\text{con}}]).
\end{aligned}
\label{eqa:cross_attention}
\end{equation}In this setup, we introduce a new linear layer, \( {L}_{\text{cross}}^q \), designed to enhance the model’s ability to attend to image-instruction features effectively. 
This additional layer has been shown to improve performance, as validated in our experiments. 
Overall, this cross-attention operation facilitates controlled image generation, leveraging desired attributes from both the conditional image and the task instruction.

\textbf{Position Embedding.}
The positional information of conditional image features is crucial for pixel-level control tasks, as methods like ControlNet incorporate features through pixel-to-pixel addition. 
To improve pixel-level control precision, we integrate Rotary Position Embedding (RoPE)~\cite{rope} into both the query and key features prior to cross-attention.
RoPE provides relative positional encoding, ensuring that query and key features with closer pixel coordinates yield higher similarity scores, thereby improving spatial alignment in the generated output.
It has been used in T2I~\cite{blackforestlabs_flux_2024} and text-to-video~\cite{yang2024cogvideox} models.

Specifically, we apply 1-D RoPE separately to the height and width dimensions. Given a feature vector \( f \in \mathbb{R}^{|D|} \) at position \( (h, w) \), we split \( f \) into two halves: \( f_h \in \mathbb{R}^{|D|/2} \) for the height dimension and \( f_w \in \mathbb{R}^{|D|/2} \) for the width. For the height component \( f_h \), the rotary matrix \( R_h \) is defined as:
\begin{small}
\setlength{\arraycolsep}{1.5pt}
\begin{equation}
R_h = 
\begin{bmatrix}
\cos (h\theta_0) & -\sin (h\theta_0) & \cdots & 0 & 0 \\
\sin (h\theta_0) & \cos (h\theta_0) & \cdots & 0 & 0 \\
\vdots & \vdots & \ddots & \vdots & \vdots \\
0 & 0 & \cdots & \cos (h\theta_l) & -\sin (h\theta_l) \\
0 & 0 & \cdots & \sin (h\theta_l) & \cos (h\theta_l) \\
\end{bmatrix},
\end{equation}
\end{small}where \( \theta_i = b^{-i/|D|} \), with \( b = 10000 \) and \( l = |D|/4 - 1 \). 
The rotary matrix \( R_w \) for the width component \( f_w \) is defined analogously. 
The updated feature vector \( f \) is then computed as:
\begin{equation}
f= [ R_h f_h \Vert R_w f_w ].
\end{equation}

RoPE is applied to both query and key features in Equation~\ref{eqa:cross_attention} and Equation~\ref{eqa:attn_ist_con}, with the position coordinates of \( K_{\text{ins}} \) set to \( (0, 0) \).
By embedding relative positional information in this way, RoPE enhances the model’s precision in pixel-level control, while maintaining robustness and adaptability across various tasks.

\section{Experiments}
\label{sec:experiment}
In this section, we conduct extensive experiments to evaluate the performance of our UNIC-Adapter in various conditional image generation tasks. 
Detailed ablation studies are also conducted to analyze the contribution of individual components within our framework. 
Additional experimental results are available in the Appendix.

\subsection{Tasks and Datasets}
Our evaluation covers three major tasks: pixel-level spatial control, subject-driven image generation, and style-image-based T2I generation. 
Below, we briefly describe the datasets used for training each task.

\textbf{Pixel-level Spatial Control.} For pixel-level spatial control, we employ the MultiGen-20M dataset~\cite{qin2023unicontrol}, which includes 12 types of pixel-level control annotations, such as Canny, HED, Sketch, Depth, Normal, Skeleton, Bbox, Segmentation, Outpainting, Inpainting, Deblurring, and Colorization. 
The dataset contains 2.8 million images with 20 million control annotations, and we reserve 1,000 samples for qualitative evaluation. 
For segmentation map conditioning, we also use the ADE20k~\cite{ade20k} training set, which includes approximately 20,000 samples. These annotations serve as conditional images, and we use GPT-4o~\cite{openai2024gpt4technicalreport} to generate 20 synonymous instructions for each control type, such as ``Generate an image from this edge map."

\textbf{Subject-driven Image Generation.} For subject-driven image generation, we use three datasets: OpenImages v7~\cite{openimages}, GRIT~\cite{kosmos-2}, and OpenStory++~\cite{ye2024openstory++}. 
Following the methodology of Subject-Diffusion~\cite{ma2024subject}, we utilize Blip-2~\cite{li2023blip} to generate captions for images without annotations and use spaCy\footnote[3]{https://github.com/explosion/spaCy} to extract subject tags. GroundingDINO~\cite{groundingdino} is used to obtain bounding boxes for each subject, and SAM~\cite{sam} is employed to generate subject segmentation maps based on these bounding boxes. 
After processing and filtering, we obtain 2.1 million subject-image pairs. 
We also use GPT-4o to generate 20 instructions, such as ``Generate an image from this dog image."

\textbf{Style-image-based Image Generation.} For style-image-based generation, we use the WikiArt~\cite{wikiart} and StyleBooth~\cite{han2024stylebooth} datasets. For StyleBooth, we randomly select an image from the same style subset as the conditional image, and for WikiArt, we employ CLIP-I~\cite{clip} similarity to retrieve a conditional image for each training image. These two datasets comprise approximately 90,000 images. Additionally, we use GPT-4o to generate 20 instructions, such as ``Generate an image based on this style image."

\begin{table}[tbp]
\centering
\begin{small}
\setlength{\tabcolsep}{2.pt}
\begin{tabular}{l|c|c|c|c}
\toprule
Method  & Canny & HED & Seg. & Depth \\
 & (F1 Score$\uparrow$) & (SSIM$\uparrow$) & (mIoU$\uparrow$) & (RMSE$\downarrow$) \\
\midrule
\multicolumn{5}{c}{Single-task} \\
\midrule
T2I-Adapter~\cite{t2iadapter}   & 23.65 & - & 12.61& 48.40 \\
Gligen~\cite{li2023gligen}  & 26.94 & 0.5634 & 23.78 & 38.83 \\
ControlNet~\cite{controlnet}  & 34.65 & 0.7621 & 32.55 & 35.90 \\
ControlNet++~\cite{controlnetplus}  & \underline{37.04} & 0.8097 & \underline{43.64} & \textbf{28.32} \\
\midrule
\multicolumn{5}{c}{Multi-task} \\
\midrule
UniControl~\cite{qin2023unicontrol}  & 30.82 & 0.7969 & 25.44 & 39.18 \\
Uni-ControlNet~\cite{zhao2024uni-controlnet}  & 27.32 & 0.6910 & 19.39 & 40.65 \\
OmniGen~\cite{xiao2024omnigen}  & 35.54 & \underline{0.8237} & \textbf{44.23} & \underline{28.54} \\
Ours & \textbf{38.94}&\textbf{0.8369}&42.89&31.10 \\
\bottomrule
\end{tabular}
\caption{Results of different methods on four pixel-level control image generation tasks. $\uparrow$ indicates that higher values are better, while $\downarrow$ indicates that lower values are better. The best results for each metric are \textbf{bolded}, and the second-best results are \underline{underlined}.}
\label{table:multigen}
\end{small}
\end{table}
\begin{table}[tbp]
\centering
\setlength{\tabcolsep}{2.5pt}
\begin{tabular}{l|ccc}
\toprule
Method & DINO$\uparrow$ & CLIP-I$\uparrow$ & CLIP-T$\uparrow$ \\
\midrule
\multicolumn{4}{c}{Fine-Tuning} \\
\midrule
Textual Inversion~\cite{mokady2023null} & 0.569 & 0.780 & 0.255 \\
DreamBooth~\cite{ruiz2023dreambooth} & 0.668 & 0.803 & \underline{0.305} \\
BLIP-Diffusion~\cite{li2024blip_diff} & 0.670 & 0.805 & 0.302 \\
\midrule
\multicolumn{4}{c}{Test Time Tuning Free} \\
\midrule
Re-Imagen~\cite{chen2022re_imagen} & 0.600 & 0.740 & 0.270 \\
SuTI~\cite{suti} & 0.741 & 0.819 & 0.304 \\
Kosmos-G~\cite{pan2023kosmos-g} & 0.694 & \textbf{0.847} & 0.287 \\
OmniGen~\cite{xiao2024omnigen} & \underline{0.801}&\textbf{0.847} & 0.301 \\
Ours & \textbf{0.816} & \underline{0.841} & \textbf{0.306} \\
\bottomrule
\end{tabular}

\caption{Results of different methods on the DreamBench dataset~\cite{ruiz2023dreambooth} for subject-driven image generation.}
\label{table:subject}
\vspace{-4mm}
\end{table}
\subsection{Implementation Details}
Our UNIC-Adapter framework is based on the SD3 medium~\cite{sd3} model. 
Since instruction texts and conditional are variations of text and image modalities, the UNIC-Adapter is initialized using the base T2I model’s parameters, which minimizes learning difficulty.
We freeze the parameters of base SD3 medium model and the feed-forward layer after the attention layer in adapter blocks, and train all other newly introduced parameters, resulting in approximately 1.2 billion trainable parameters.
We use the AdamW optimizer~\cite{adamw} with a learning rate of 0.0001 and a weight decay of 0.01. Training is performed on 16 H100 GPUs for 100,000 steps, with a batch size of 16 per GPU.

\begin{figure*}[t]
  \centering
   \includegraphics[width=0.99\linewidth]{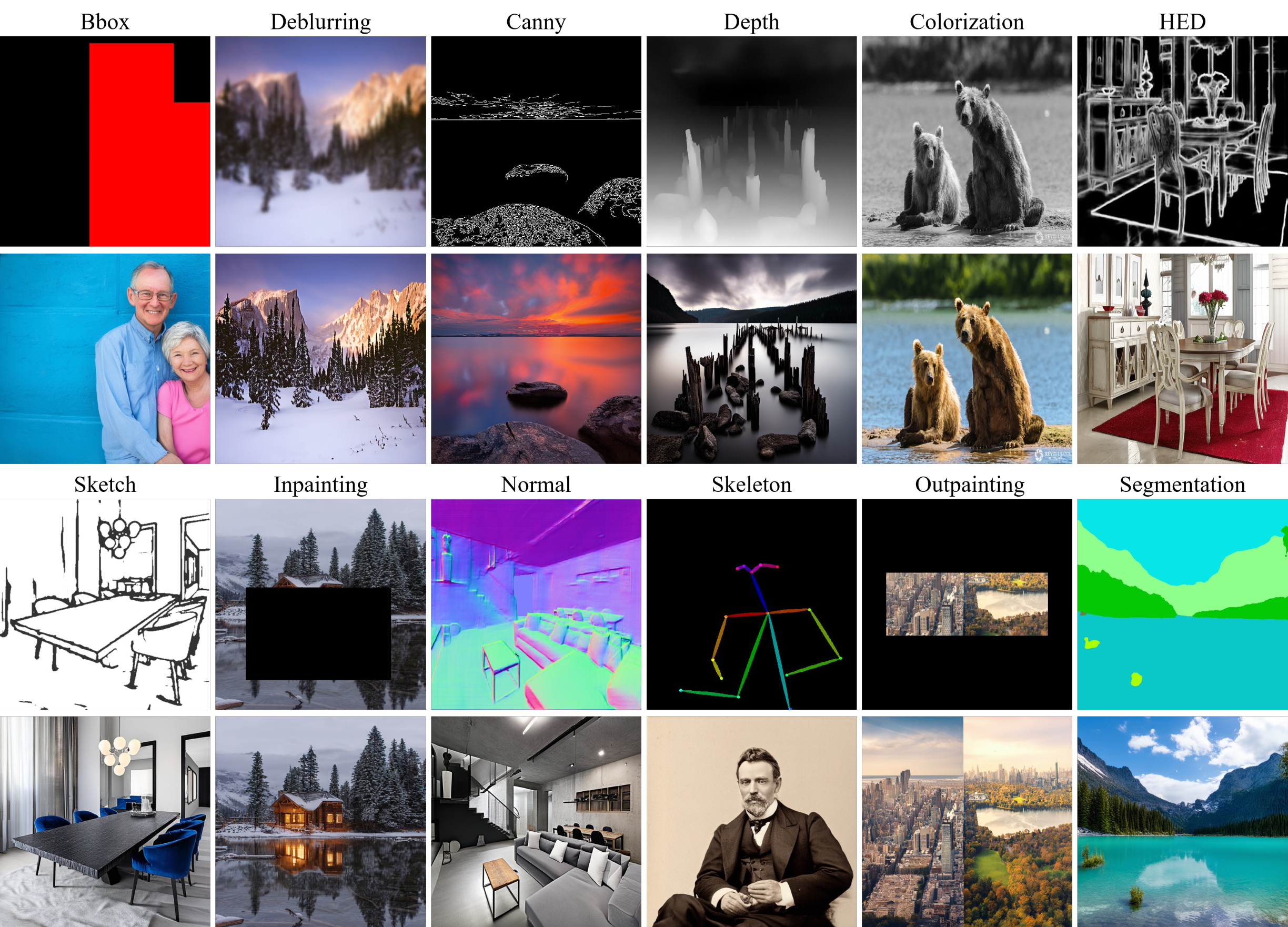}
   \caption{Visualization results of our UNIC-Adapter on twelve pixel-level control tasks from the MultiGen-20M dataset. The first and third rows show different types of conditional images, while the second and fourth rows display the corresponding generated images.}
   \label{fig:pixel}
\end{figure*}

\subsection{Main Results}
We present experimental results of the proposed UNIC-Adapter framework across three major tasks.

\textbf{Pixel-Level Spatial Control.} Following previous work~\cite{controlnetplus,xiao2024omnigen}, we evaluate performance based on the similarity between control conditions extracted from generated images and the specified pixel-level control inputs. 
The evaluation is conducted on the test set of MultiGen-20M with canny edge, hed edge, and depth map conditions and on ADE20k test set with segmentation map condition. 
As shown in Table~\ref{table:multigen}, UNIC-Adapter achieves the best performance on canny and hed edge map conditions, and demonstrates comparable performance on depth and segmentation map conditions, compared to SOTA methods like ControlNet++~\cite{controlnetplus} for single-task and OmniGen~\cite{xiao2024omnigen} for multi-task scenarios. 
Figure~\ref{fig:pixel} shows qualitative results on 12 tasks from MultiGen-20M, illustrating that UNIC-Adapter consistently produces images accurately aligned with pixel-level controls.

\begin{figure}[t]
  \centering
   \includegraphics[width=0.99\linewidth]{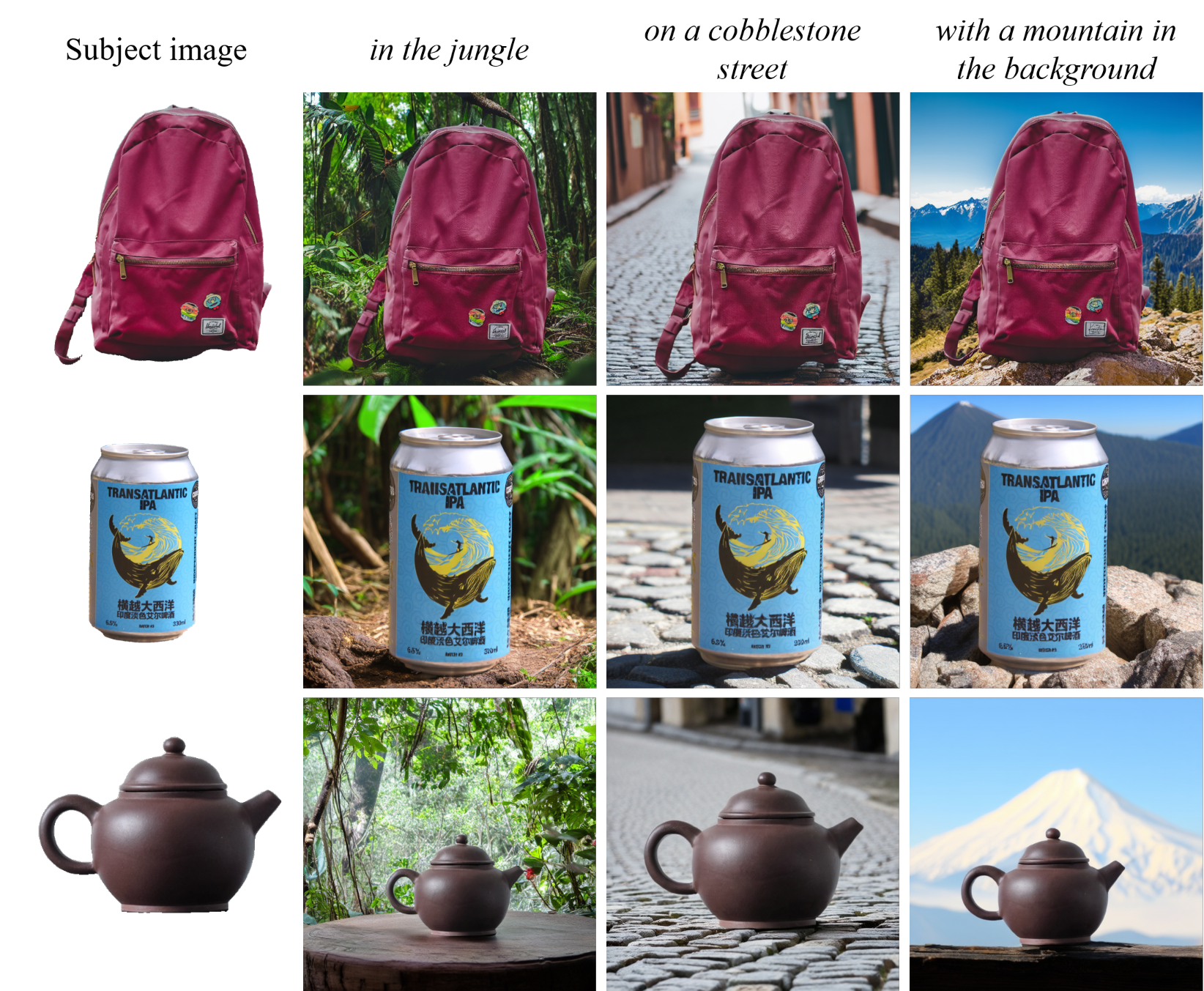}
   \caption{Visualization results of our UNIC-Adapter on DreamBench for subject-driven generation. The first column displays the subject images, while the other three columns show the generated images based on different prompts.}
   \label{fig:subject}
\end{figure}

\textbf{Subject-Driven Image Generation.} Evaluation is conducted using 30 objects and 25 prompts from the DreamBench dataset~\cite{ruiz2023dreambooth}. Following prior work~\cite{pan2023kosmos-g}, we generate 4 images for each prompt and measure subject fidelity using DINO~\cite{dino} and CLIP-I scores and text fidelity with CLIP-T~\cite{clip}. 
As shown in Table~\ref{table:subject}, UNIC-Adapter achieves high subject-image consistency and maintains strong alignment between the generated images and text prompts. Qualitative examples in Figure~\ref{fig:subject} demonstrate that UNIC-Adapter effectively generates images closely resembling input subjects across various contexts.

\begin{figure}[t]
  \centering
   \includegraphics[width=0.99\linewidth]{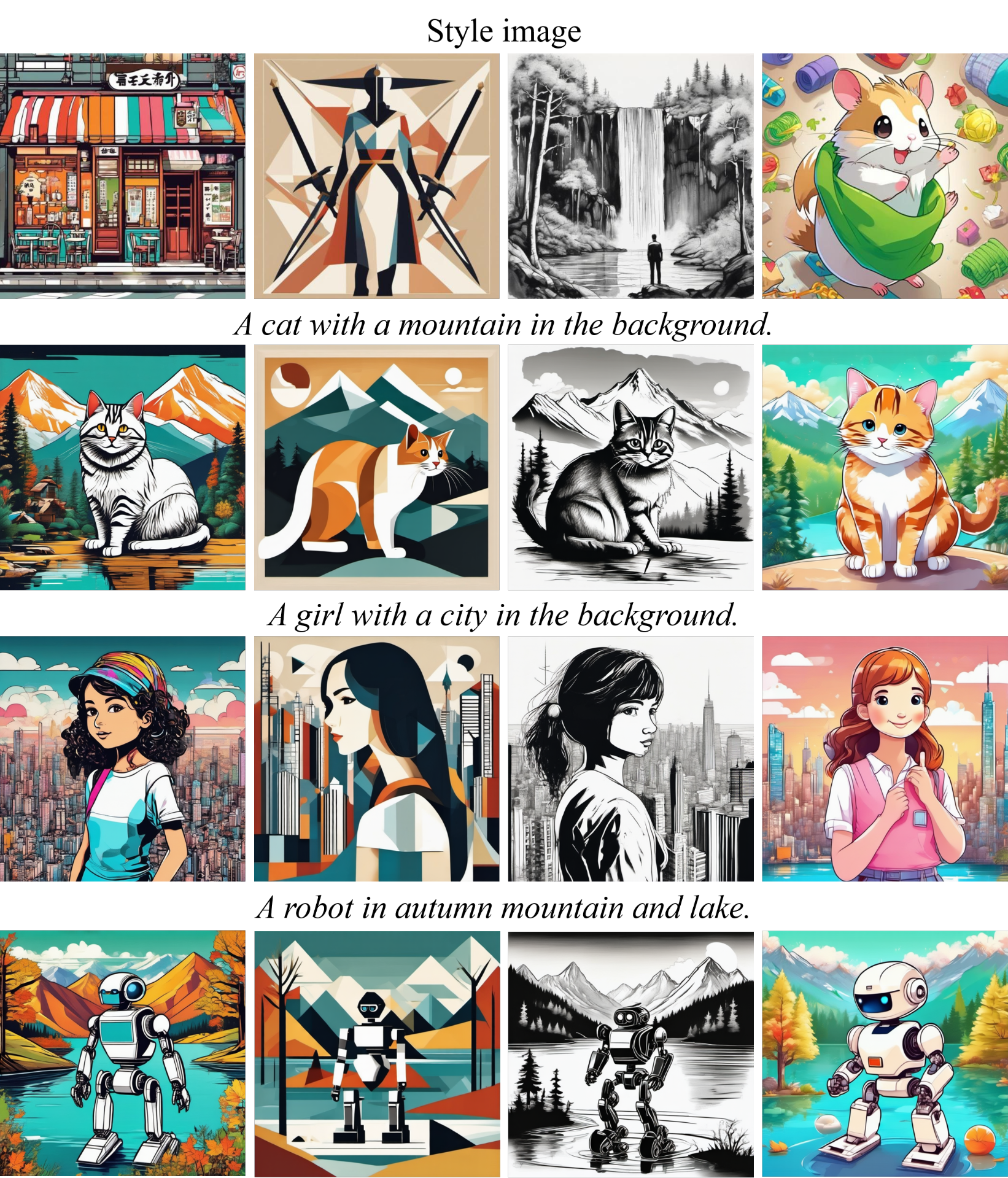}
   \caption{Visualization results of our UNIC-Adapter on style-image-based T2I generation. The first row shows the reference style image, and each subsequent row contains images generated from the same prompt, influenced by different style images.}
   \label{fig:style}
   \vspace{-4mm}
\end{figure}

\textbf{Style-image-based Image Generation.} For the same text prompt, style images from StyleBooth~\cite{han2024stylebooth} are randomly selected as conditional images. The qualitative results in Figure~\ref{fig:style} indicate that UNIC-Adapter effectively captures and applies stylistic elements from conditional images, generating artistically consistent outputs.


\subsection{Ablation Studies}
To better understand the impact of individual components in the UNIC-Adapter framework, we conduct ablation studies by systematically modifying or removing key modules. For efficiency, the models in this part are trained on the subject-driven image generation task and four tasks (canny, hed, depth, and segmentation) from MultiGen-20M.

\textbf{Query and Key Feature Choices.}
We investigate the effect of different query and key feature choices in Equation~\ref{eqa:cross_attention}.
First, we fix the query features as \( Q_{\text{txt}} \) and \( Q_{\text{img}} \) and examine the impact of different key features. 
As shown in Table~\ref{table:ablation_query_key}, \( K_{\text{con}} \) plays a critical role in pixel-level spatial control tasks and in maintaining subject-image similarity. 
Relying solely on \( K_{\text{ist}} \) as key feature (Exp. 3) significantly reduces performance across pixel-level control tasks and subject-image similarity. 
We also investigate the impact of query features and find that using \( Q_{\text{txt}} \) alone as the query feature  (Exp. 5) also degrades performance. 
In general, using \( Q_{\text{img}} \) as the query and both \( K_{\text{con}} \) and \( K_{\text{ist}} \) as key features (Exp. 4) yields optimal performance on pixel-level control tasks and comparable performance on subject-driven generation tasks.
This highlights the importance of task instruction features in unified controllable generation. We adopt this configuration for subsequent experiments.

\begin{figure}[t]
  \centering
   \includegraphics[width=0.99\linewidth]{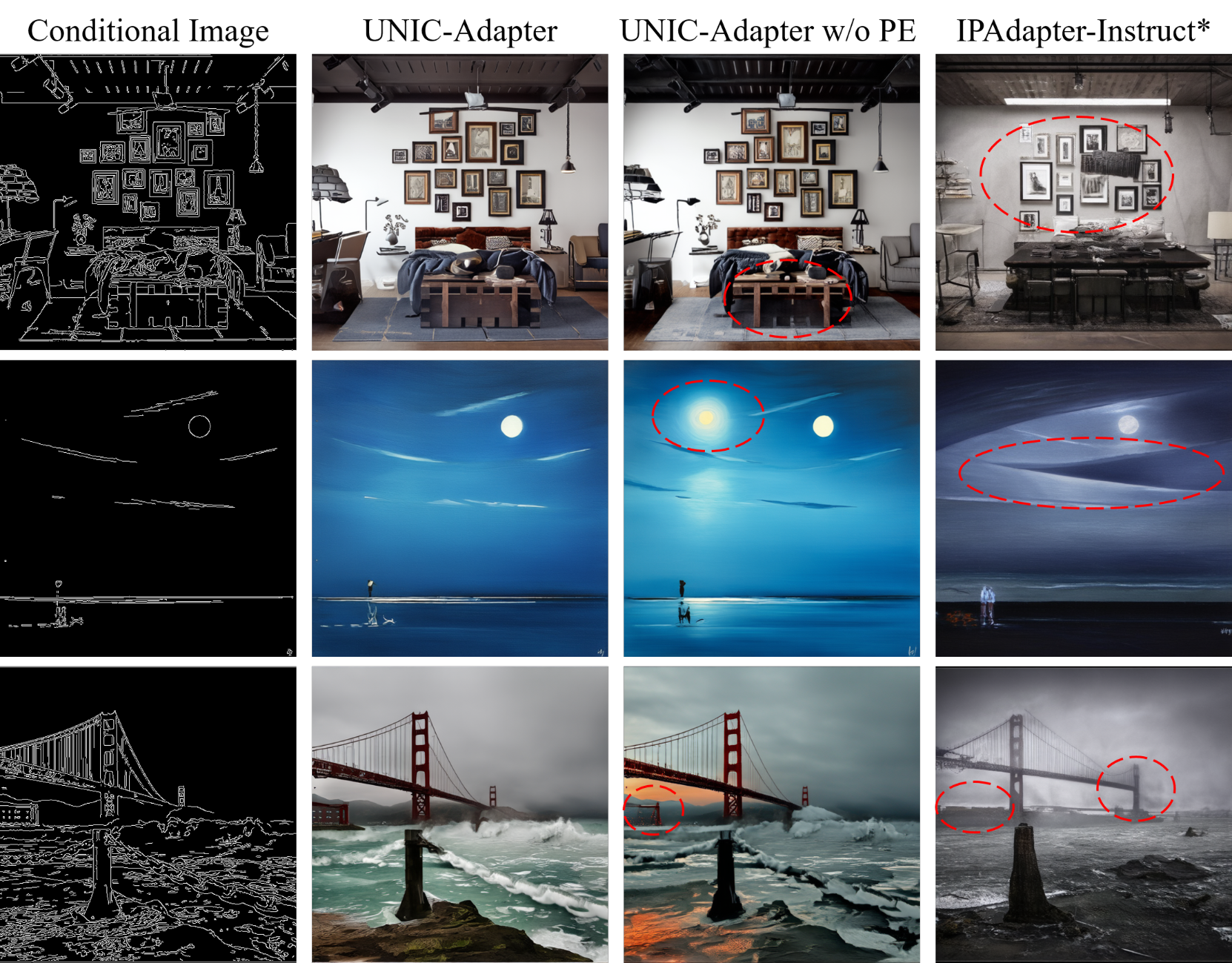}
   \caption{Comparison of visualization results of our UNIC-Adapter with and without position embedding, and IPAdapter-Instruct* on pixel-level control generation using canny edge conditional images. UNIC-Adapter generates images that are more closely aligned with the conditional images compared to the other two methods.}
   \label{fig:ablation_pe}
\end{figure}
\begin{figure}[t]
  \centering
   \includegraphics[width=0.99\linewidth]{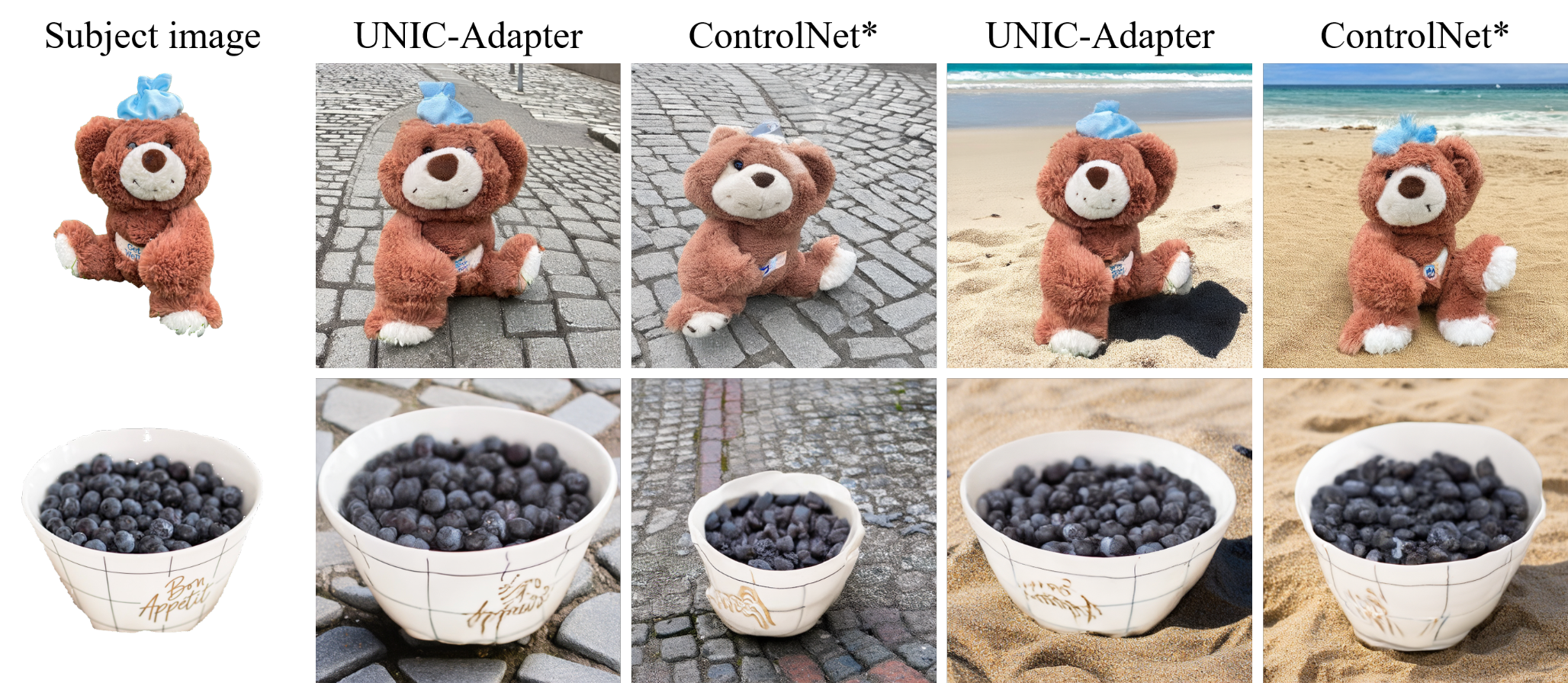}
   \caption{Comparison of visualization results between our UNIC-Adapter and ControlNet* on subject-driven generation. Compared to ControlNet*, UNIC-Adapter achieves better consistency between subject images and generated images.}
   \label{fig:ablation_baseline}
   \vspace{-2mm}
\end{figure}
\begin{table*}
\centering
\setlength{\tabcolsep}{3.75pt}
\begin{tabular}{c|cc|cc | cccc |ccc}
\toprule
Exp. &\multicolumn{2}{c|}{Query} & \multicolumn{2}{c|}{Key} & Canny & HED & Seg. & Depth & \multicolumn{3}{c}{Subject-Driven} \\
Number &$Q_{img}$ & $Q_{txt}$ & $K_{con}$ & $K_{ist}$ & (F1 Score$\uparrow$) & (SSIM$\uparrow$) &(mIoU$\uparrow$)  & (RMSE$\downarrow$) &  DINO$\uparrow$ & CLIP-I$\uparrow$ & CLIP-T$\uparrow$  \\
\midrule
1 &\checkmark & \checkmark & \checkmark & \checkmark & \underline{29.98} &\underline{0.7840}&\underline{30.84}&\underline{34.05} & \textbf{0.778}&\textbf{0.827}&0.309  \\
2 &\checkmark & \checkmark & \checkmark &  & 29.01&0.7767&29.00&35.39  &  \underline{0.769}&\underline{0.825}&0.307  \\
3 &\checkmark & \checkmark &  & \checkmark & 22.38&0.5599&22.96&38.82  &  0.694&0.780&\textbf{0.319}  \\
\midrule
4 &\checkmark &  & \checkmark & \checkmark & \textbf{31.32}&\textbf{0.7934}&\textbf{30.94}&\textbf{33.16}  &  \underline{0.769}&0.823&0.308  \\
5&& \checkmark & \checkmark & \checkmark & 19.10&0.3730&17.06&46.95  &  0.669&0.782&\underline{0.317}  \\
6&\checkmark &  & \checkmark &  & 29.24&0.7786&27.81&34.12  &  0.750&0.817&0.313  \\
\bottomrule
\end{tabular}
\caption{Results of different query and key feature choices on pixel-level control tasks and subject-driven generation task.}
\label{table:ablation_query_key}
\end{table*}

\begin{table*}
\centering
\begin{tabular}{l | cccc |ccc}
\toprule
Method & Canny & HED & Seg.  & Depth & \multicolumn{3}{c}{Subject-Driven} \\ 
       & (F1 Score$\uparrow$) &  (SSIM$\uparrow$)& (mIoU$\uparrow$) & (RMSE$\downarrow$) &  DINO$\uparrow$ & CLIP-I$\uparrow$ & CLIP-T$\uparrow$  \\

\midrule
w/o PE & 31.32&0.7934&30.94&33.16 & 0.769&0.823&0.308  \\
w/ Abs PE& 31.51&0.7967&30.31&32.98 & 0.740&0.809&\textbf{0.314}  \\
w/ Abs PE + ${L}_{\text{cross}}^q$& 36.93&0.8348&33.22&32.74 & 0.751&0.811&\textbf{0.314}  \\
w/ RoPE & 34.65&0.8248&32.13&33.45 & 0.761&0.817&\underline{0.313}  \\
w/ RoPE + ${L}_{\text{cross}}^q$ & \underline{37.95}&\textbf{0.8420}&\underline{33.32}&\underline{32.25} & \underline{0.784}&\underline{0.829}&0.309  \\
w/o txt & 36.31 & 0.8387 & 33.71 & 31.60 & 0.788&0.830&0.308  \\
w/o txt update & 36.31 & 0.8387 & 33.71 & 31.60 & 0.772&0.821&0.310  \\
\midrule
ControlNet* & \textbf{38.18}&\underline{0.8382}&\textbf{34.81}&\textbf{31.46} & 0.749&0.815&\underline{0.313}  \\
IPAdapter-Instruct* & 15.81&0.2191&16.12&47.81 & \textbf{0.788}&\textbf{0.852}&0.294  \\

\bottomrule
\end{tabular}
\caption{Results of two baseline methods and different module combinations of our UNIC-Adapter on pixel-level control tasks and subject-driven generation task.}
\label{table:ablation_pe}
\vspace{-2mm}
\end{table*}

\textbf{Importance of Position Embedding.}
To enhance the control precision of pixel-level tasks, we experiment with incorporating different types of position embeddings, including the absolute position embedding in SD3 and the RoPE used in our adapter, to query and key features of each layer.
As shown in Table~\ref{table:ablation_pe}, RoPE yields better performance than absolute embeddings.
We further examine the impact of introducing a new linear layer, \( {L}_{\text{cross}}^q \), for query image features in the main branch. 
Table~\ref{table:ablation_pe} shows that the addition of \( {L}_{\text{cross}}^q \) improves results further, particularly in pixel-level control tasks and subject-image similarity.
This indicates the importance of \( {L}_{\text{cross}}^q \) in enabling query features \( Q_{\text{img}} \) to retrieve specific features from conditional image and task instruction. 
Figure~\ref{fig:ablation_pe} shows that UNIC-Adapter without position embedding often generates images with incomplete or incorrect details from control signals.

\textbf{Comparison with Baseline Methods.}
We reproduce two baseline methods, ControlNet* and IPAdapter-Instruct*, in our framework. 
For ControlNet*, we replace the cross-attention mechanism in the image-instruction feature injection module with an addition operation (as did in ControlNet) while retaining other settings. 
For IPAdapter-Instruct*, we reproduce results by referring to the open-source code\footnote[4]{https://github.com/unity-research/IP-Adapter-Instruct} on our datasets.
As shown in Table \ref{table:ablation_pe}, ControlNet* achieves performance comparable to our final model on pixel-level control tasks but exhibits suboptimal subject-image similarity on the subject-driven generation task. 
The visualization results in Figure~\ref{fig:ablation_baseline} show that ControlNet* often produces images with distorted subject content.
In contrast, IPAdapter-Instruct* performs well on subject-driven generation tasks but poorly on pixel-level control tasks. 
Due to compressing conditional images into 16 tokens without precise spatial information, the generated images from IPAdapter-Instruct* in Figure~\ref{fig:ablation_pe} fail to accurately capture the pixel-level details of the conditioned images.
In comparison, our UNIC-Adapter achieves consistently superior results across both task types, demonstrating its effectiveness and suitability for unified controllable image generation.
\section{Conclusion}
In this paper, we present UNIC-Adapter, a unified framework for controllable T2I generation that integrates diverse conditional image inputs and task instructions within a single model. 
Built on the MM-DiT architecture, our approach leverages cross-attention mechanisms and RoPE to achieve precise pixel-level control, as well as high subject fidelity, across multiple generation tasks.
Our extensive experiments on tasks such as pixel-level spatial control, subject-driven image generation, and style-image-based T2I generation demonstrate the effectiveness of the UNIC-Adapter. 

\clearpage
{
    \small
    \bibliographystyle{ieeenat_fullname}
    \bibliography{main}
}

\clearpage
\setcounter{page}{1}
\maketitlesupplementary

\section{Additional implementation details}
In this section, we provide additional implementation details of our UNIC-Adapter.

\subsection{Model Architecture}
As described in the main paper, our UNIC-Adapter shares the same architecture as the SD3 medium model~\cite{sd3} and is initialized using the parameters of the SD3 medium.
Specifically, our UNIC-Adapter consists of 24 MM-DiT blocks, with each block containing two AdaLayerNormZero layers, one Attention layer, and two Feed-Forward layers.
To reduce the number of trainable parameters, we freeze the parameters of the Feed-Forward layers and only train the remaining layers.

\subsection{Training Details}
The dataset mixing ratios are set as follows: pixel-level spatial control: 0.4, subject-driven image generation: 0.5, and style-image-based image generation: 0.1.
For subject-driven image generation, the background of the subject images is set to white.
The input images are first resized so that the shorter side is 512 pixels, and then they are randomly cropped to a resolution of \(512 \times 512\). 
To enable classifier-free guidance, for pixel-level spatial control, we use a probability of 0.15 to drop the text prompt.
For the other two tasks, we use the following probabilities: 0.05 to drop the text prompt, 0.05 to drop both the task instruction and the conditional image simultaneously, and 0.05 to drop the text prompt, task instruction, and conditional image simultaneously.
Our UNIC-Adapter is trained using the same loss function as SD3 medium~\cite{sd3}. 

\subsection{Inference Details}
During inference, we use the same sampling schedule as SD3, with the sampling step set to 28. 
We employ classifier-free guidance based on three conditions: the text prompt $c_{txt}$, the task instruction $c_{ist}$, and the conditional image $c_{con}$.
The classifier-free guidance is performed as follows:
\begin{equation}
\begin{aligned}
& e_{\theta}(z_{t}, c_{txt}, c_{ist}, c_{con}) = e_{\theta}(z_{t}, \varnothing, \varnothing, \varnothing) \\
& + s_{c} \cdot (e_{\theta}(z_{t}, \varnothing, c_{ist}, c_{con}) -  e_{\theta}(z_{t}, \varnothing, \varnothing, \varnothing)) \\
& + s_{t} \cdot (e_{\theta}(z_{t}, c_{txt}, c_{ist}, c_{con}) - e_{\theta}(z_{t}, \varnothing, c_{ist}, c_{con}))
\end{aligned}
\end{equation}where $e_{\theta}$ denotes the model, $z_{t}$ denotes the image latents, $\varnothing$ denotes the fixed null value, $s_{c}$ is the scale for image-instruction guidance, and $s_{t}$ is the scale for text prompt guidance.
For pixel-level spatial control, $s_{c}$ and $s_{t}$ are set to 1.3 and 3.0, respectively.
For subject-driven image generation, $s_{c}$ and $s_{t}$ are set to 1.2 and 7.5, respectively.
For style-image-based generation, $s_{c}$ and $s_{t}$ are set to 3.0 and 6.0, respectively.

\section{Additional Experimental Results}
In this section, we present additional experimental results, including both quantitative ablation studies and qualitative evaluations.

\subsection{Importance of Cross-modal Interaction}
Our UNIC-Adapter leverages the MM-DiT block, where task instruction features and conditional image features mutually attend to each other. 
To investigate the importance of this cross-modal interaction, we perform an experiment by modifying the attention process in Equation (4) of the main paper to align with the DiT block formulation~\cite{chen2023pixart}. 
The modified process is defined as follows:
\begin{small}
\begin{equation}
\begin{aligned}
& K_{\text{ist}} = {L}_{\text{ist}}^k(Z_{\text{ist}}), V_{\text{ist}} = {L}_{\text{ist}}^v(Z_{\text{ist}}), \\
& Q_{\text{con}} = {L}_{\text{con}}^q(Z_{\text{con}}), K_{\text{con}} = {L}_{\text{con}}^k(Z_{\text{con}}), V_{\text{con}} = {L}_{\text{con}}^v(Z_{\text{con}}), \\
& Z_{\text{con}}^{'} = \text{Attn}(Q_{\text{con}}, [K_{\text{ist}} \Vert K_{\text{con}}], [V_{\text{ist}} \Vert V_{\text{con}}]),
\end{aligned}
\end{equation}
\end{small}where the task instruction features no longer attend to conditional image features and instead act solely as key and value features without being updated. 
As shown in Table~\ref{table:cross_modal}, removing cross-modal interaction leads to a decline in performance across several metrics, emphasizing the advantage of such interaction between task instruction features and conditional image features.

\subsection{More Qualitative Evaluation}
\begin{figure*}[t]
  \centering
   \includegraphics[width=0.90\linewidth]{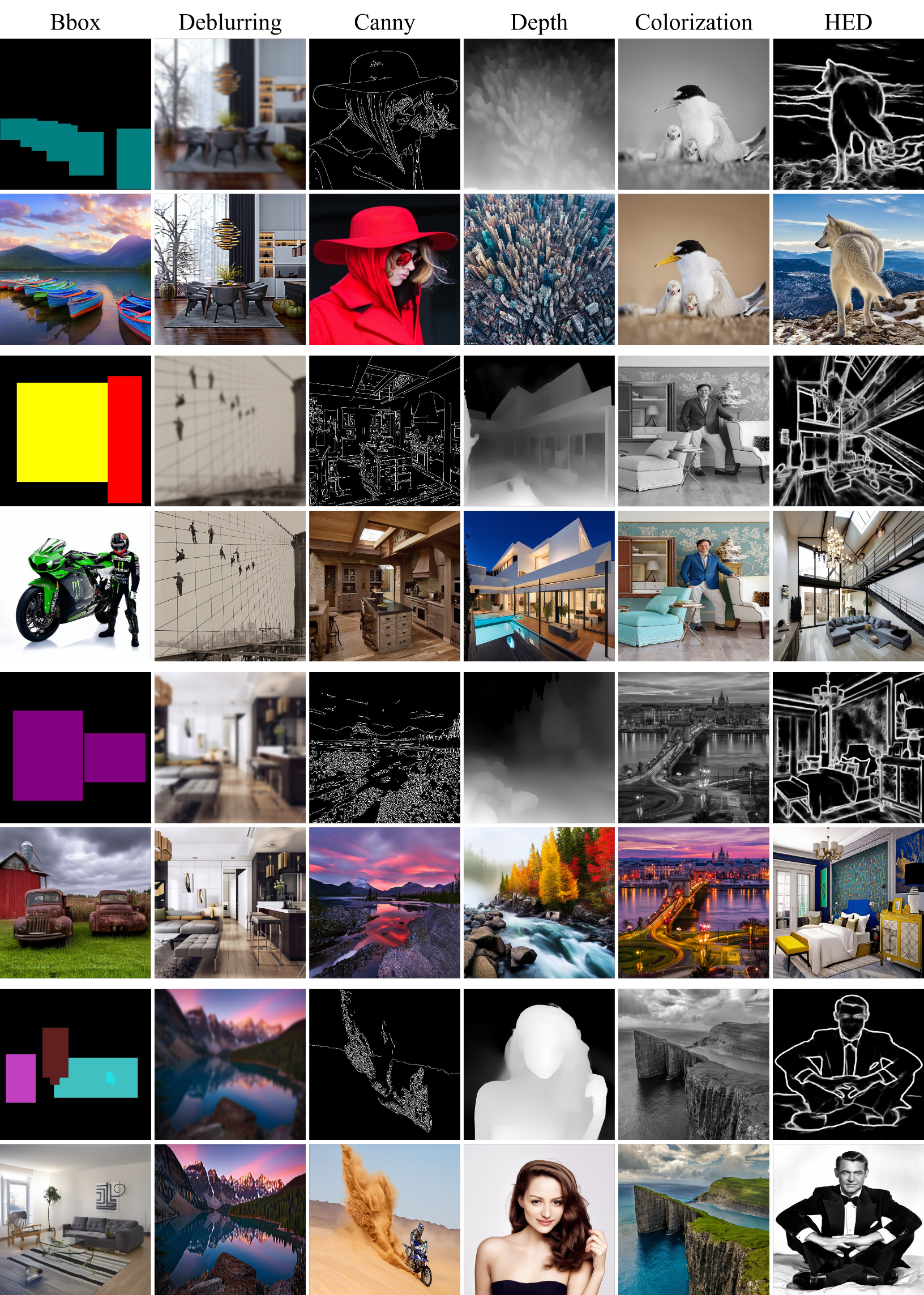}
   \caption{Visualization results of our UNIC-Adapter on six pixel-level control tasks from the MultiGen-20M dataset. The odd rows show different types of conditional images, while the even rows display the corresponding generated images.}
   \label{fig:pixel_1}
\end{figure*}

\begin{figure*}[t]
  \centering
   \includegraphics[width=0.90\linewidth]{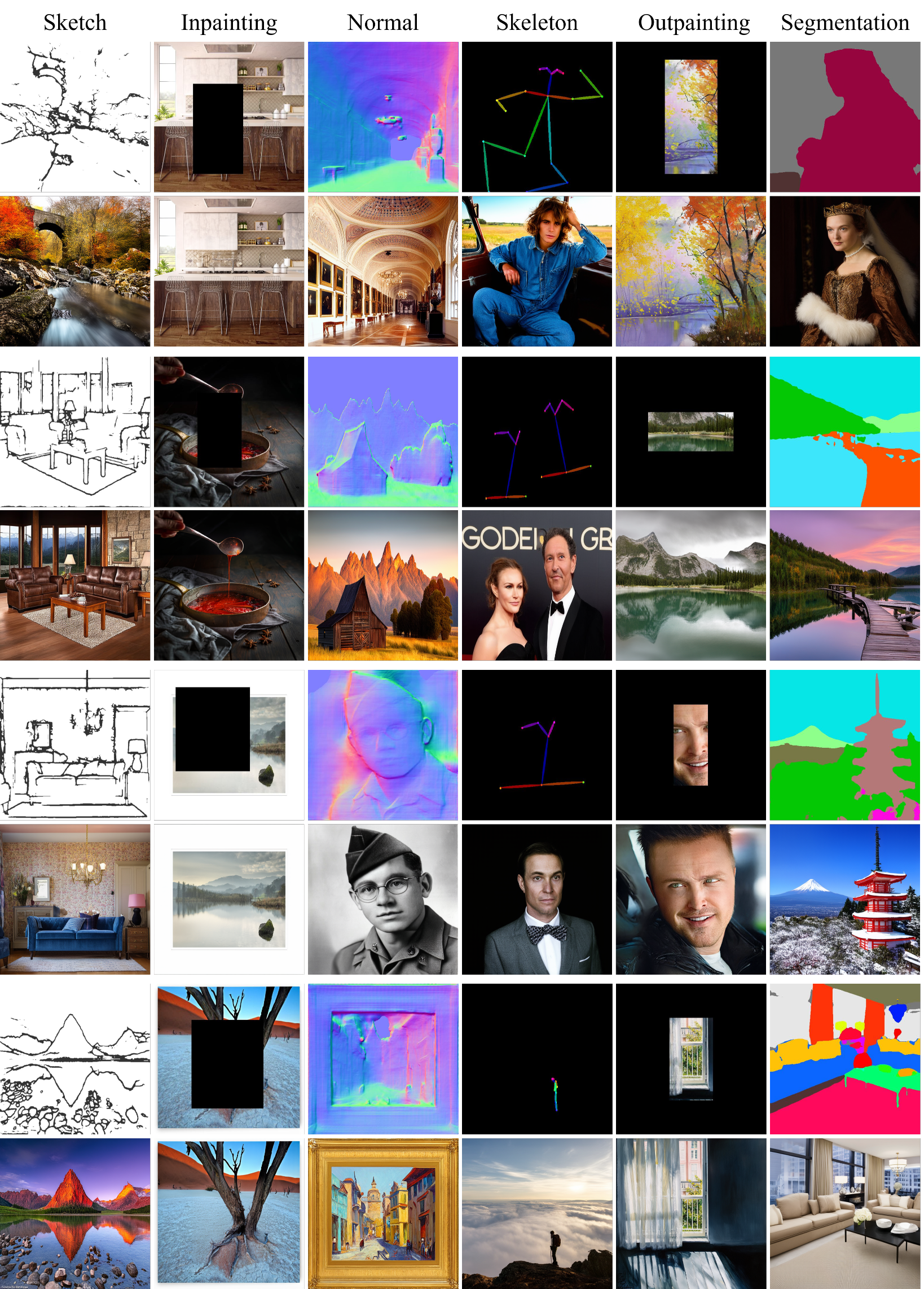}
   \caption{Visualization results of our UNIC-Adapter on six pixel-level control tasks from the MultiGen-20M dataset. The odd rows show different types of conditional images, while the even rows display the corresponding generated images.}
   \label{fig:pixel_2}
\end{figure*}

\begin{figure*}[t]
  \centering
   \includegraphics[width=0.90\linewidth]{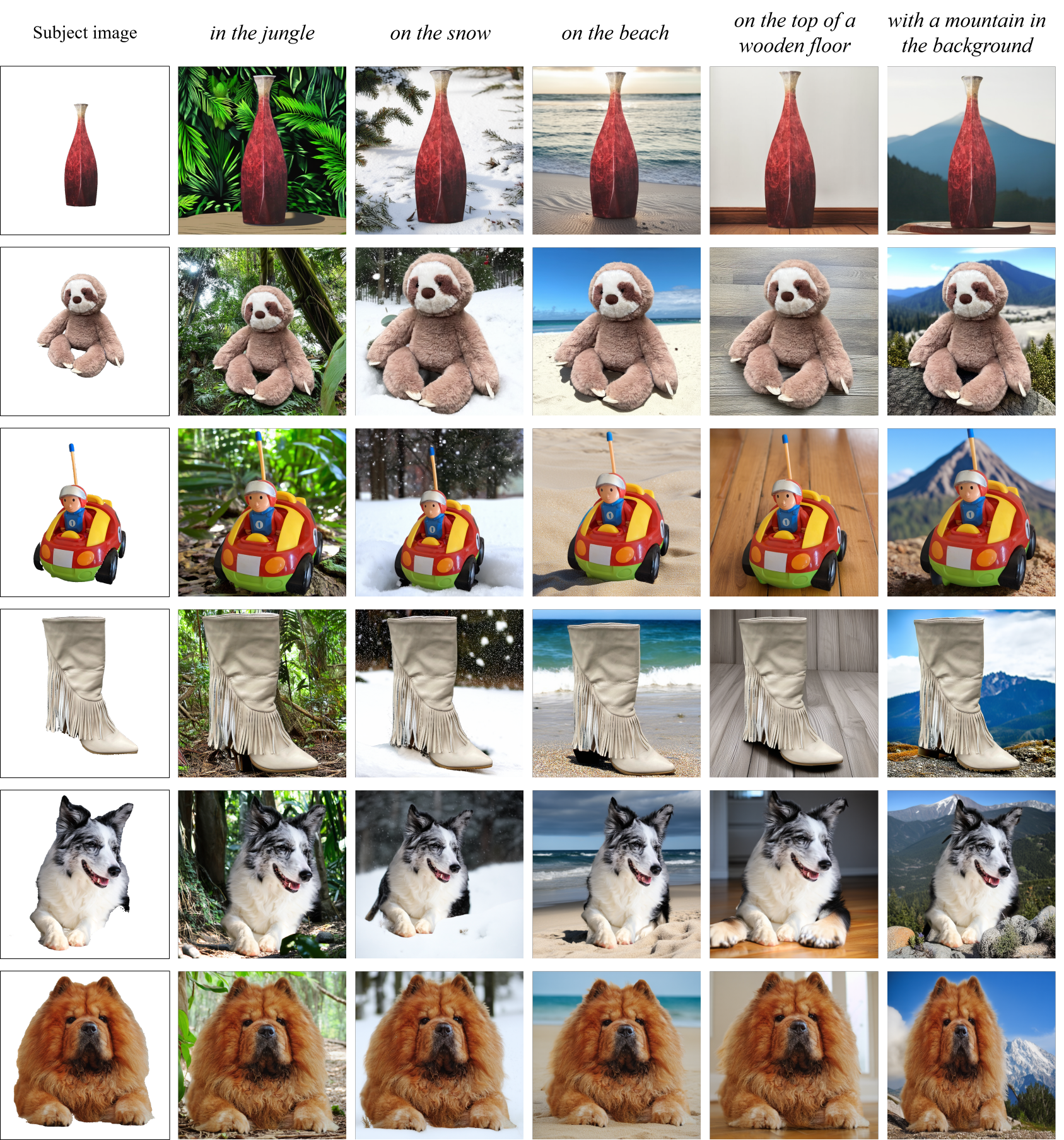}
   \caption{Visualization results of our UNIC-Adapter on DreamBench for subject-driven generation. The first column displays the subject images, while the other columns show the generated images based on different prompts.}
   \label{fig:subject_1}
\end{figure*}

\begin{figure*}[t]
  \centering
   \includegraphics[width=0.90\linewidth]{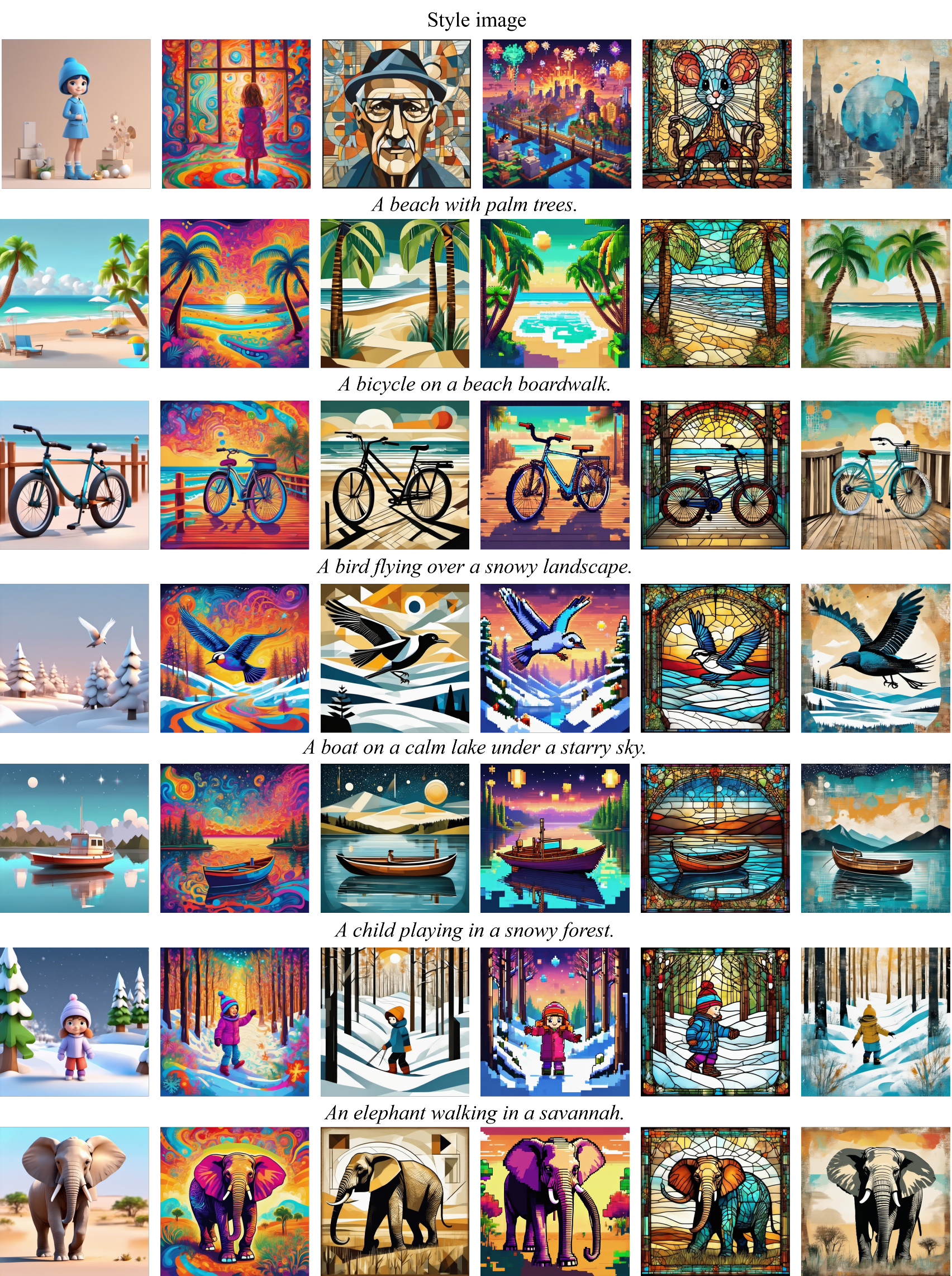}
   \caption{Visualization results of our UNIC-Adapter on style-image-based T2I generation. The first row shows the reference style images, and each subsequent row contains images generated from the same prompt, influenced by different style images.}
   \label{fig:style_1}
\end{figure*}

Figures~\ref{fig:pixel_1},~\ref{fig:pixel_2},~\ref{fig:subject_1}, and~\ref{fig:style_1} provide additional visualization results of our UNIC-Adapter across various controllable generation tasks.

\begin{table*}[t]
\centering
\begin{tabular}{l | cccc |ccc}
\toprule
Method & Canny & HED & Seg.  & Depth & \multicolumn{3}{c}{Subject-Driven} \\ 
       & (F1 Score$\uparrow$) &  (SSIM$\uparrow$)& (mIoU$\uparrow$) & (RMSE$\downarrow$) &  DINO$\uparrow$ & CLIP-I$\uparrow$ & CLIP-T$\uparrow$  \\

\midrule
UNIC-Adapter & 37.95&0.8420&33.32&32.25 & 0.784&0.829&0.309  \\
w/o cross-modal & 37.71 & 0.8284 & 31.73 & 31.95 & 0.772&0.821&0.310  \\

\bottomrule
\end{tabular}
\caption{Results of UNIC-Adapter with and without cross-modal interaction between task instruction features and conditional image features on pixel-level control tasks and subject-driven generation task.}
\label{table:cross_modal}
\end{table*}

\section{Limitations and Future Work}
Due to the limitation of image resolution size in the training datasets we used, for convenience, all training images are resized and cropped to a resolution of \(512 \times 512\) in our experiments.
As a result, our UNIC-Adapter is limited in generating images with higher resolution, \(1024 \times 1024\).
In future work, we will collect more high-resolution training images, such as images with pixel areas equivalent to \(1024 \times 1024\).
Additionally, in the subject-driven image generation task, the generated subjects exhibit limited variations in pose compared to the subject images, since the subject image and target image originate from the same image source during training.
To address this limitation, future efforts should include incorporating video data or leveraging data synthesis techniques to generate training samples.
As demonstrated in prior works~~\cite{ms-diffusion,xiao2024omnigen}, utilizing subject and target images from different sources proves effective in generating diverse subject images across varying contexts.
Finally, integrating our UNIC-Adapter with state-of-the-art T2I models, such as FLUX1.0-dev~\cite{blackforestlabs_flux_2024} and Stable Diffusion 3.5 Large~\cite{sd3}, might further enhance the controllability and performance of these models.

\end{document}